\documentclass{article}
\usepackage{spconf,amsmath,epsfig}
\usepackage{bm}
\usepackage[numbers,sort&compress]{natbib}
\usepackage{color}
\usepackage{appendix}

\let\OLDthebibliography\thebibliography
\renewcommand\thebibliography[1]{
  \OLDthebibliography{#1}
  \setlength{\parskip}{0pt}
  \setlength{\itemsep}{0pt plus 0.3ex}
}

\begin{document}\sloppy

% Example definitions.
% --------------------
\def\x{{\mathbf x}}
\def\L{{\cal L}}

% Title.
% ------
\title{2D-guided 3D Gaussian Segmentation}
%
% Single address.
% ---------------
%Address and e-mail should NOT be added in the submission paper. They should be present only in the camera ready paper. 
\address{Kun Lan\textsuperscript{1}, Haoran Li\textsuperscript{1}, Haolin Shi\textsuperscript{1}, Wenjun Wu\textsuperscript{1}, Yong Liao\textsuperscript{1}, Lin Wang\textsuperscript{2}, Pengyuan Zhou*\textsuperscript{1}\\
\textsuperscript{1}University of Science and Technology of China,\textsuperscript{2}AI Thrust, HKUST(GZ)\\
\tt \small\{lankun,lhr123,mar,wu\_wen\_jun,yliao\}@mail.ustc.edu.cn \\
\tt \small linwang@ust.hk, zpymyyn@gmail.com
}
% \author{Kun Lan\textsuperscript{1}, Haoran Li\textsuperscript{1}, Haolin Shi\textsuperscript{1}, Wenjun Wu\textsuperscript{1}, Yong Liao\textsuperscript{1}, Lin Wang\textsuperscript{2}, Pengyuan Zhou*\textsuperscript{1}\\
% \textsuperscript{1}University of Science and Technology of China, \textsuperscript{2}AI Thrust, HKUST(GZ)\\
% {\tt\small \{lankun,lhr123,mar,wu_wen_jun,yliao\}@mail.ustc.edu.cn} \\
% {\tt\small linwang@ust.hk, zpymyyn@gmail.com  }
% }

\maketitle
\begin{abstract}
Recently, 3D Gaussian, as an explicit 3D representation method, has demonstrated strong competitiveness over NeRF (Neural Radiance Fields) in terms of expressing complex scenes and training duration. These advantages signal a wide range of applications for 3D Gaussians in 3D understanding and editing. Meanwhile, the segmentation of 3D Gaussians is still in its infancy. The existing segmentation methods are not only cumbersome but also incapable of segmenting multiple objects simultaneously in a short amount of time. In response, this paper introduces a 3D Gaussian segmentation method implemented with 2D segmentation as supervision. This approach uses input 2D segmentation maps to guide the learning of the added 3D Gaussian semantic information, while nearest neighbor clustering and statistical filtering refine the segmentation results. Experiments show that our concise method can achieve comparable performances on mIOU and mAcc for multi-object segmentation as previous single-object segmentation methods.
\end{abstract}
\section{Introduction}
\label{sec:intro}
% 最近兴起的3D高斯技术比起诸如点云、mesh、符号距离函数、神经辐射场等3D表征有着明显的优势，尤其是在训练时间与场景重建质量上。它使用高斯分布的均值代表其在空间中的位置，协方差矩阵代表旋转与尺寸，球谐函数来表达颜色。以SFM得到的点云为起始，3D高斯天生就包含了场景的几何信息，这让他省去了在空间中寻找物体集中区域的时间，而显式的表达方式又进一步省去了为空间中每个点计算颜色与密度的时间，使得其可以做到实时渲染。此外，自适应密度控制赋予了其表达细节特征的能力。这些优势使得其在3D理解与编辑中有着广泛的应用前景，但作为这些应用所必须的3D高斯分割却少有研究。
The recently emerged 3D Gaussian technique~\cite{3D-Gaussian} marks a significant advancement over previous 3D representation methods such as point clouds~\cite{point-cloud}, meshes~\cite{mesh}, signed distance functions (SDF)~\cite{SDF}, and neural radiance fields (NeRF)~\cite{NeRF}, especially in terms of training time and scene reconstruction quality. The mean of each 3D Gaussian represents the position of its center point, the covariance matrix indicates rotation and size, and spherical harmonics express color. Starting with point clouds obtained from SFM~\cite{SFM}, 3D Gaussians inherently contain the scene's geometric information, thus saving time in locating areas with concentrated objects in space. Moreover, their explicit expression method further accelerates calculations of color and density for every 3D Gaussian in space, enabling real-time rendering. Additionally, adaptive density control endows them with the capability to express detailed features. These advantages make it widely applicable in 3D understanding and editing. Nonetheless, there is little research on 3D Gaussian segmentation, which is another critical pillar of the realm.

% 最近提出的高斯分割方法Gaussian Grouping需要较长的训练时间，SAGA实现较为繁琐且难以同时对多物体进行分割。而3D高斯的显式表达带来的存储开销又使得其不能像NeRf分割一样直接将2D语义特征学习到3D中。最后，数据集的稀少以及缺乏标注使得其无法像2D分割与点云分割一样采用有监督的方式实现分割。
A few Gaussian segmentation methods have been proposed recently, yet they require further improvement. For example, Gaussian Grouping~\cite{Gaussian_Grouping} requires an extended training period of about 15 minutes. SAGA~\cite{SAGA} is complex in its implementation and struggles with segmenting multiple objects simultaneously. Additionally, the explicit expression of 3D Gaussians leads to storage overhead, preventing it from directly transferring 2D semantic features into 3D, as in NeRF segmentation~\cite{Decomposing,ISRF}. Finally, the scarcity of datasets and the lack of annotations impede the application of supervised segmentation methods, commonly utilized in 2D and point cloud segmentation.

% 基于以上问题以及受2D分割为每个像素分配一个其在各类别上的概率分布向量启发，我们使用一个预训练的2D分割模型来指导3D高斯分割。首先，我们为每个3D高斯附加一个object code，长度为K的概率向量，用来表示每个高斯在不同类别上的概率分布。接着，我们使用算法，通过最小化2D分割图和此位姿下渲染后分割图的误差来引导每个高斯的分类。具体说来，我们将渲染后2D分割图的像素类别认为是在渲染时当前射线上多个高斯类别的加权。我们认为第一个高斯对其贡献最大，此后每个高斯的贡献按其与渲染平面间的距离依次递减。同时该权重也与高斯的大小成正比。最后，在实验中我们发现仅使用以上算法会导致一些高斯的语义信息不准确以及某些高斯被错误的分割出来。为此我们采用KNN聚类来解决高斯语义模糊的问题，可选的统计过滤也可以去除那些被错误分割的高斯。在以物体为中心和360°场景中的实验证明了本方法的有效性。
In light of the aforementioned challenges, we propose leveraging a pre-trained 2D segmentation model to guide 3D Gaussian segmentation. Inspired by the 2D segmentation approach, which assigns a probability distribution vector for each pixel across different categories, we first assign an object code to each 3D Gaussian to indicate the Gaussian's categorical probability distribution. Subsequently, we employ an algorithm that guides the classification of each 3D Gaussian by minimizing the error between the 2D segmentation map and the rendered segmentation map at a given pose. 
%Specifically, we consider the pixel categories in the rendered 2D segmentation map as a weighted sum of the categories of multiple Gaussians along the current ray during rendering. We assume the first Gaussian contributes the most, with each subsequent Gaussian's contribution diminishing in accordance with its distance from the rendering plane, and this weight is also proportional to the size of the Gaussian. 
Finally, %in our experiments, we observed that just using the aforementioned algorithm resulted in some Gaussians having inaccurate semantic information and several Gaussians being incorrectly segmented. To address this, 
we employ KNN clustering to resolve semantic ambiguity in 3D Gaussians and statistical filtering to remove erroneously segmented 3D Gaussians. We validated the effectiveness of our approach through experiments in object-centric and 360° scenes. Our contributions can be summarized as follows.
\begin{itemize}
  \item We propose an efficient 3D Gaussian segmentation method supervised by 2D segmentation, which can learn the semantic information of a 3D scene in less than two minutes and segment multiple objects in 1-2 seconds for a given viewpoint.
  \item Extensive experiments on LLFF, NeRF-360, and Mip-NeRF 360 have demonstrated the effectiveness of our method, obtaining an mIOU of 86$\%$.
  %\item Our method will have a wide range of applications in downstream tasks such as 3D understanding and editing.
\end{itemize}
\section{Related Work}
% 三维高斯表示作为最近提出的一种显式表示方法，在三维场景重建中取得了显著的成就，且可以实现场景的实时渲染。给定利用一系列场景图像和相应的相机数据，Gaussian Splatting使用3D高斯来描绘场景对象。每个高斯由包括均值、协方差矩阵、不透明度和球面谐波在内的参数定义。平均值确定位高斯在3D场景中的中心位置。用缩放矩阵S和旋转矩阵R表示的协方差矩阵描述高斯的大小和形状，而球面谐波对其颜色信息进行编码。然后，Gaussian Splatting使用基于点的渲染进行高效的3D到2D投影。
\textbf{3D Gaussian}, a recently proposed explicit representation method, has attained remarkable achievements in three-dimensional scene reconstruction~\cite{3D-Gaussian}. Its biggest advantage is the capability of real-time rendering. Utilizing a series of scene images and corresponding camera data, it employs 3D Gaussians to depict scene objects. Each 3D Gaussian is defined by parameters including mean, covariance matrix, opacity, and spherical harmonics. The mean pinpoints the Gaussian's central position in the 3D scene. Expressed by a scaling matrix $S$ and a rotation matrix $R$, the covariance matrix describes the Gaussian's size and shape, while the spherical harmonics encode its color information. Gaussian Splatting then utilizes point-based rendering for efficient 3D to 2D projection.

% 最近3D高斯产生了许多的延申工作。DreamGaussian和GaussianDreamer等将该技术与Diffusion模型融合，促进了文本到3D的生成。4D高斯飞溅将这些方法扩展到动态场景表示和渲染。在分割方面，Gaussian Grouping和SAGA取得了显著进展。两者都使用SAM来导出2D先验分割数据，指导3D高斯中语义信息的学习。在Gaussian Grouping中，这种信息类似于球面谐波函数系数来传递，而SAGA使用可学习的低维特征。然而，SAM对几何结构的依赖限制了其在每个掩码中的语义包容性。因此，这两种方法都提出了保证不同视角下SAM分割结果一致性的策略。Gaussian Grouping将来自不同角度的图像视为视频帧序列，利用预先训练的模型进行掩模传播和匹配。SAGA采用定制的SAM-guidance loss来聚合跨视角一致的多粒度分割信息。
Recent developments have seen numerous advancements in Gaussian Splatting. Innovations like DreamGaussian~\cite{DreanGaussian} and GaussianDreamer~\cite{GaussianDreamer} merge this technique with Diffusion model~\cite{Diffusion}, facilitating text-to-3D generation. 4D Gaussian Splatting~\cite{4D} extends these methods to dynamic scene representation and rendering. Focusing on segmentation, Gaussian Grouping~\cite{Gaussian_Grouping} and SAGA~\cite{SAGA} have made significant strides. They both employ the Segment Anything Model (SAM)~\cite{SAM} to derive 2D prior segmentation data, guiding the learning of added semantic information in 3D Gaussians. In Gaussian Grouping, this information is conveyed similarly to coefficients of spherical harmonic functions, whereas SAGA uses learnable low-dimensional features. However, SAM's reliance on geometric structures limits its semantic inclusivity in each mask. Thus, both methods propose strategies to ensure consistency of SAM's segmentation outcomes from various perspectives. Gaussian Grouping treats images from different angles as a sequence of video frames, utilizing a pre-trained model for mask propagation and matching. In contrast, SAGA consolidates consistent, multi-granularity segmentation information across viewpoints, employing a custom-designed SAM-guidance loss.

% 作为此前流行的一种3D表示方法，NeRF产生了大量的延申工作，其中不乏一些对NeRF进行分解和分割的工作。Object-NeRF，提出一个dual-pathway的神经辐射场。其场景分支处理空间坐标和观看方向，从观看者的角度输出点的密度和颜色细节，主要对3D场景的背景进行编码，并为对象分支提供几何上下文。独特的是，除了空间和方向输入外，对象分支还集成了可学习的对象激活码，从而能够独立学习每个场景对象的神经辐射场。提出的3D guard mask可以缓解物体之间的遮挡问题。类似地，Switch NeRF通过可训练的门控网络来实现大规模神经辐射场的分解。
\noindent\textbf{3D Segmentation in Radiance Fields.} Prior to the advent of 3D Gaussians, NeRF~\cite{NeRF} stood as a prominent method in 3D characterization, sparking a plethora of derivative works~\cite{DM-NeRF,ISRF,Object-NeRF,OR-NeRF,SegNeRF,SPIn-NeRF,Switch-NeRF}, including several focusing on decomposing and segmenting NeRF. A notable example is Object NeRF~\cite{Object-NeRF}, which introduced a dual-pathway neural radiance field adept at object decomposition. Its scene branch processes spatial coordinates and viewing directions, outputting density and color details of a point from the viewer's perspective, primarily encoding the background of the 3D scene and offering geometric context for the object branch. Uniquely, the object branch, in addition to spatial and directional inputs, integrates a learnable object activation code, enabling the independent learning of neural radiance fields for each scene object. And the 3D guard msak helps mitigate occlusion issues between objects during the learning phase. Similarly, Switch-NeRF~\cite{Switch-NeRF} demonstrates the decomposition of large-scale neural radiance fields through a trainable gating network. 

% DM-NeRF引入了一个对象场来实现NeRF分割，使用它来生成一个指示对象对每个空间点的所有权的单热向量。SPIn-NeRF采用语义辐射场，评估场景位置与特定对象相关联的可能性。ISRF将语义特征添加到特定点，并通过师生模型将渲染图像的DINO特征合并到该框架中。任意点的特征通过插值得到。融合了K-means聚类、最近邻匹配和双边搜索等技术，实现了交互式NeRF分割。此外，OR-NeRF选择将2D分割结果反向投影到3D空间中，将它们传播到不同的视角，然后将它们重新渲染到2D平面上。这些方法或是需要较长的时间，或是难以在分割结果中保留场景的细节特征。为此，我们提出在短时间内完成对场景中多物体的分割并保留细节特征。
DM-NeRF~\cite{DM-NeRF} introduces an object field for NeRF segmentation, using it to generate a one-hot vector indicating the ownership of each spatial point by an object. SPIn-NeRF~\cite{SPIn-NeRF} employs a semantic radiance field, assessing the likelihood of scene locations being associated with specific objects. ISRF~\cite{ISRF} adds semantic features to specific points and incorporates DINO~\cite{DINO} features of rendered images into this framework through a teacher-student model, allowing for feature interpolation at any given point. Techniques such as K-means clustering, nearest neighbor matching, and bilateral search are integrated, enabling interactive NeRF segmentation. Additionally, OR-NeRF\cite{OR-NeRF} chooses to back-project 2D segmentation results into a 3D space, propagating them across different viewpoints, and then re-rendering them onto a 2D plane. 

These 3D Gaussian and NeRF segmentation methods either take a long time or struggle to preserve the detailed features of the scene in the segmentation result. For this reason, we propose a method that can segment multiple objects while preserving the detailed features in a short time.
\begin{figure*}[ht]
    \centering
    \includegraphics[scale=0.6]{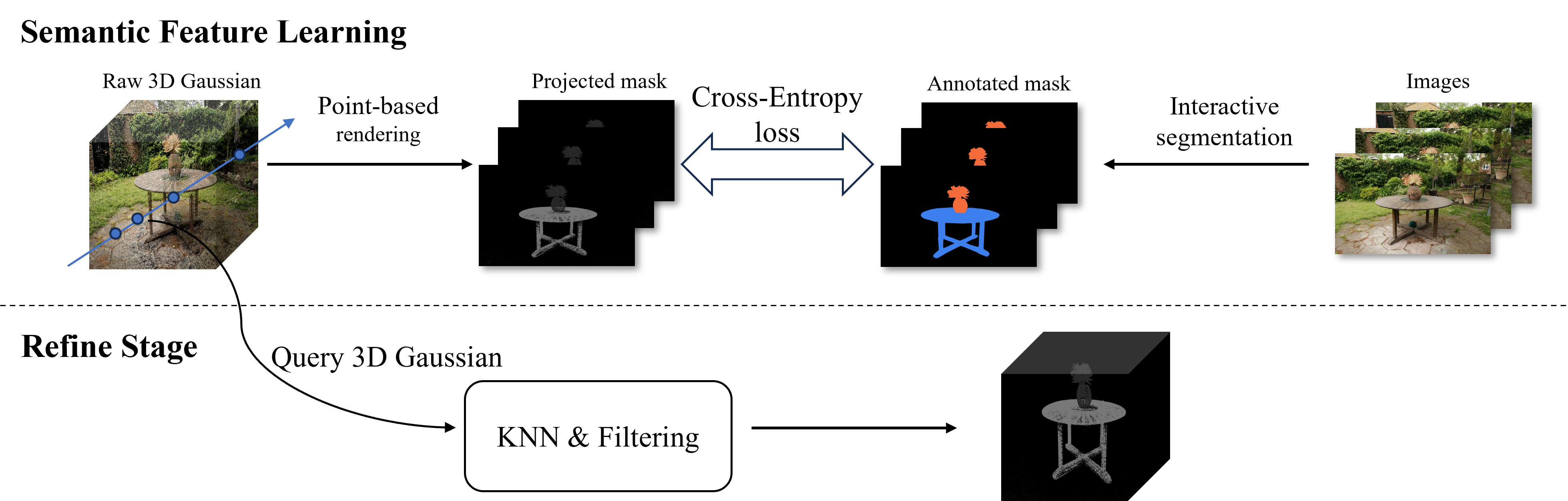}
    \caption{The pipeline of our method. Given posed training images, we first utilize an interactive model to get the 2D prior segmentation information, then add semantic information to the 3D Gaussians, and project this information onto a 2D plane to make loss with the 2D prior knowledge, finally we use KNN and statistical filtering to refine the segmentation result.}
    \label{pipeline}
\end{figure*}
\section{Method}
% 给定一个训练好的使用3D高斯表示的场景，场景渲染图与这些渲染图对应的相机参数。我们首先使用一个交互式的2D分割模型来分割这些渲染图。接着，得到的分割图被用作指导来帮助3D高斯添加的语义信息的学习。最后，我们采用KNN聚类来解决某些高斯语义信息模糊的问题，同时可选的统计过滤可以帮主去除那些被错误分割的高斯。
Given a well-trained scene using 3D Gaussian representation, scene rendering images, and corresponding camera parameters, we initially employed an interactive 2D segmentation model~\cite{eiseg} to segment the rendered images. Then, the obtained 2D segmentation maps are used as guidance to facilitate the learning of semantic information (object code) added to the 3D Gaussians. Finally, we use KNN clustering to address issues of semantic ambiguity in certain 3D Gaussians, while optional statistical filtering can help eliminate those 3D Gaussians that have been erroneously segmented. The pipeline is depicted in Fig. \ref{pipeline}.

\subsection{Point-Based rendering and Semantic Information Learning}

Gaussian Splatting~\cite{3D-Gaussian} employs a point-based rendering technique ($\alpha$-blending) to render a 3D scene onto a plane, and the color of a pixel on the plane can be calculated as:
\begin{equation} \label{alpha-blending}
    C = \sum_{i\in\mathcal{N}}c_i\alpha_i\prod_{j=1}^{i-1}(1-\alpha_j),
\end{equation}
where $\mathcal{N}$ denotes the ordered Gaussians overlapping the pixel, $c_i$ represents the color of each 3D Gaussian projected onto the current pixel, and $\alpha_i$ is given by evaluating a 2D Gaussian with covariance $\Sigma$ multiplied with a learned per-Gaussian opacity. It is worth noting that $\alpha$ expresses the opacity of any point in the projected 2D Gaussian, which decreases as its distance from the 2D Gaussian center increases.

% 为了实现对3D场景的分割，需要将语义信息添加到3D场景中。受2D分割的启发，我们为场景中的每个高斯都添加一个object code，用来表示其在不同类别上的概率分布。值得注意的是，我们定义了一个背景类，并使用object code的第一个维度来表示他。
To achieve segmentation of a 3D scene, semantic information needs to be incorporated into the representation of the scene. Inspired by 2D segmentation, we assign an object code $\bm{o}\in\mathcal{R}^K$ to each 3D Gaussian to represent the probability distribution of the current 3D Gaussian across various categories, where $K$ is the number of categories. Note that we defined a background class and the first dimension of $\bm{o}$ is used to represent it.

% 为了使用2D分割图作监督来实现添加的3D语义信息的学习，需要将添加的语义信息从3D投影到2D中。受到alpha-blending的启发，我们将渲染后2D分割图的像素类别认为是在渲染时当前射线上多个高斯类别的加权。我们认为第一个高斯对其贡献最大，此后每个高斯的贡献按其与渲染平面间的距离依次递减。同时该权重也与高斯的大小成正比。渲染图上每个像素的类别可用3D高斯的object code表示为：
To use 2D segmentation maps as supervision for learning the added 3D semantic information, it is necessary to project the added semantic information from 3D onto a 2D plane. Inspired by $\alpha$-blending, we consider the pixel categories in the rendered 2D segmentation map as a weighted sum of the categories of multiple 3D Gaussians along the current ray during rendering. We assume that the first 3D Gaussian contributes the most, with each subsequent 3D Gaussian's contribution diminishing in accordance with its distance from the rendering plane, and this contribution is also proportional to the size of the 3D Gaussian itself. The category of each pixel on the rendered image can be represented by the object code of $\bm{o}$ the 3D Gaussians as:
\begin{equation}
    \bm{\hat{o}} = \sum_{i\in\mathcal{N}}o_i\alpha_i\prod_{j=1}^{i-1}(1-\alpha_j),
\end{equation}
which simply replaces the color $c$ of each 3D Gaussian in Eq. (\ref{alpha-blending}) with the object code of each 3D Gaussian.

Assuming we have $L$ images of 2D ground truth labels $\left\{I_1,\cdots, I_l,\cdots, I_L\right\}, I_l\in\mathcal{R}^{H \times W}$, $L$ is number of different camera poses in the dataset, $H$ and $W$ are the height and width of the label respectively. Each element in the ground truth label represents the category label of the corresponding pixel. Then we generate $L$ corresponding projected segmentation maps $\left\{\Bar{I_1},\cdots, \Bar{I_l},\cdots, \Bar{I_L}\right\}, \Bar{I_l}\in\mathcal{R}^{K \times H \times W}$ in the same camera viewpoint as the ground truth. In these projected segmentation maps, each element represents the probability of the pixel belonging to $i^{th}$ category, $i=1,2,\cdots,K$.

Next, the original 2D segmentation maps are transformed into one-hot vector and then reshaped to be $M\in\mathcal{R}^{K \times N}$, where $N=H \times W$. As the projected segmentation maps, we perform a similar operation and obtain $\Bar{M}\in\mathcal{R}^{K \times N}$. Then the ground truth object mask $M$ and corresponding projected object mask $\Bar{M}$ are used to calculate the Cross-Entropy Loss(CES):
\begin{equation}
    L_{i}=-\frac{1}{N}\sum_{n=1}^{N}{M}_i^n log\Bar{M}_i^n,\ (0\le i < K).
\end{equation}
The final loss is the average of all the losses for the $L$ pairs of images:
\begin{equation}
    \mathcal{L}=\frac{1}{L}\sum_{l=1}^{L}CES_l,\ where\ CES_l=\frac{1}{K}\sum_{i=1}^{K}L_{i}.
\end{equation}

\begin{figure*}[ht]
    \centering
    \includegraphics[scale=0.5]{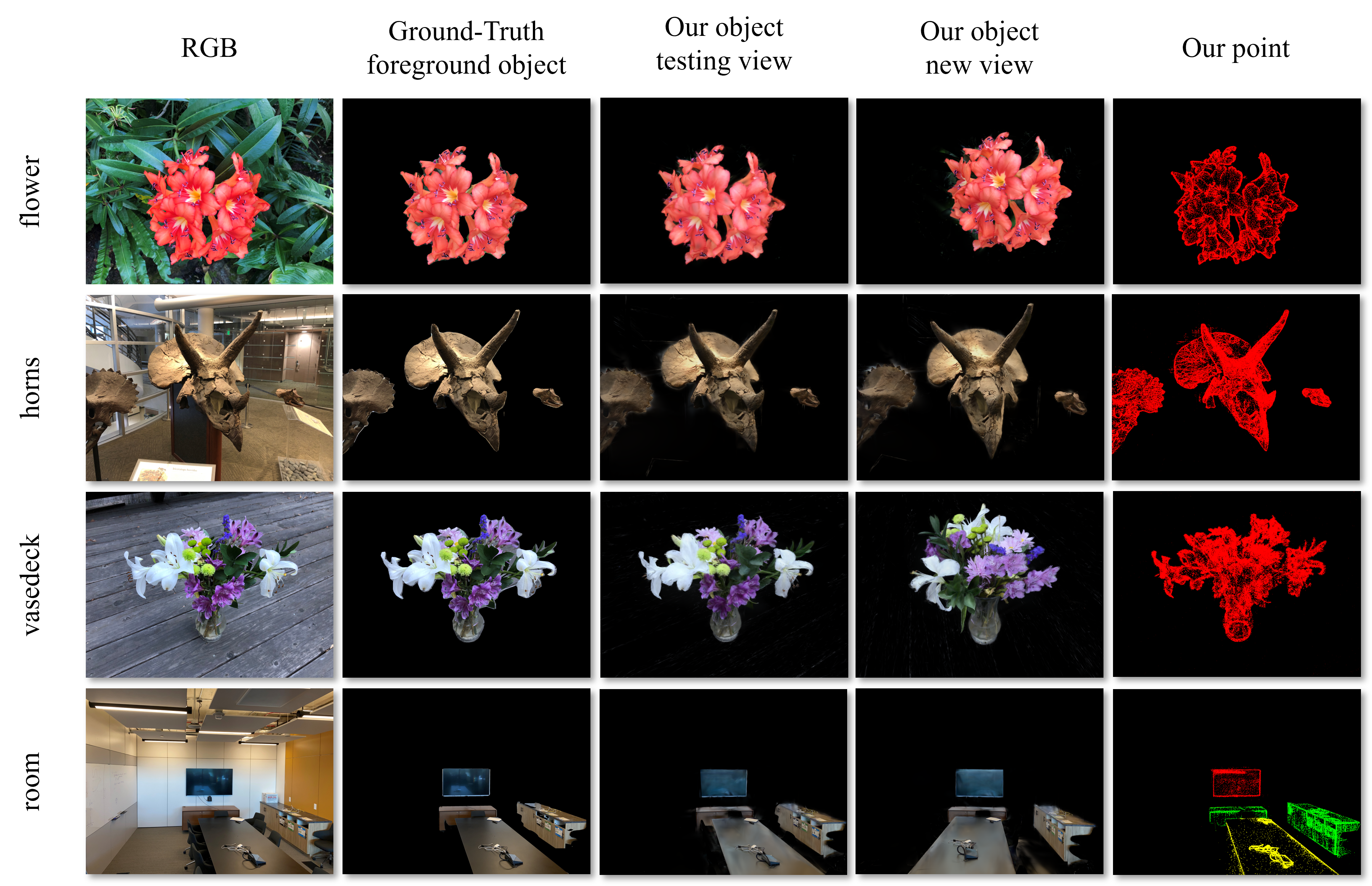}
    \caption{The qualitative results of our method. The first and second columns are the original image and the foreground object obtained from the interactive segmentation model, respectively. The third and fourth columns are renderings of the 3D segmentation effect obtained by our method from the test viewpoint and a new viewpoint. The last column is the result of converting the different categories belonging to the 3D Gaussian into RGB values.}
    \label{Experiment}
\end{figure*}

\subsection{Gaussian Clustering}
% 实验表明仅使用2D分割图作监督学习3D语义信息会使得某些3D高斯的语义信息不准确。这表现为他们在所有类别上呈现初始状态的均匀分布或者在几个类别上概率相近。为了解决这个问题，同时考虑到物体在空间中是连续分布的，我们认为每个3D高斯应当与其周围一定距离内的3D高斯是同一类别。
During experiments, we observed that employing 2D segmentation maps as the sole guide for learning 3D semantic information may lead to inaccuracies in the semantic information of some 3D Gaussians. These inaccuracies manifest either as 3D Gaussians approximating an initial state of uniform distribution across all categories or as exhibiting similar probabilities in a limited number of categories. To address this issue, and considering that objects are continuously distributed in space, we posit that each 3D Gaussian should typically be classified within the same category as other 3D Gaussians located within a certain proximity. 

% 为了解决语义信息的不准确，我们采用KNN聚类算法。对于一个带有语义信息的3D场景，我们首先获取场景中每个3D高斯的object code。接着这些object code经过softmax得到每个3D高斯在不同类别上的概率分布，那些概率最大值小于一定阈值的3D高斯会被筛选出来。最后，将这些筛选出来的3D高斯的object code和中心点坐标送入KNN算法。对于一个查询3D高斯，我们计算其与周围3D高斯的距离，K个最近邻的3D高斯被选择，查询3D高斯的object code被设置为这些筛选出来的3D高斯的object code的均值。
To remedy the inaccuracies in semantic information, we refer to the KNN clustering algorithm. For a 3D scene with pre-learned semantic information, we initially retrieve the object code, denoted as $\bm{o}$, of each 3D Gaussian used to represent the scene. These codes then undergo softmax processing to deduce the probability distribution of each 3D Gaussian across various categories. 3D Gaussians with maximum probability values $max(softmax(\bm{o}))<\beta$ are selected. Finally, we fed the object codes of these selected 3D Gaussians along with their center coordinates into KNN for clustering. For a query 3D Gaussian, we calculate its distance from the surrounding 3D Gaussians, and the $k$ 3D Gaussians closest in distance are selected, the object code of the query Gaussian is set to the mean of these 3D Gaussians' object code.

\subsection{Gaussian Filtering}
% 实验表明在经过3D语义信息学习和高斯聚类后，一些不属于待分割物体的3D高斯会被错误的分割出来。我们观察到这些3D高斯通常距离分割出的其他3D高斯较远。因此，我们使用类似于点云分割中的统计滤波算法来解决这个问题。对于每个分割出来的3D高斯，我们计算其与周围3D高斯的平均距离。接着再计算这些平均距离的均值和方差。最后，我们在分割结果中去除那些平均距离大于均值与方差和的3D高斯。
During experiments, We also found that after 3D semantic information learning and Gaussian clustering, some 3D Gaussians not belonging to the object intended for segmentation were incorrectly segmented out. We observed that these erroneously segmented 3D Gaussians are spatially distant from the rest of the segmented 3D Gaussians, as shown in Fig. \ref{Ablation}(a). Therefore, we employ a statistical filtering algorithm similar to that used in point cloud segmentation to solve this problem. For each segmented Gaussian, we calculate its average distance $D$ from the neighboring 3D Gaussians. Then, we compute the mean $\mu$ and variance $\sigma$ of these average distances. Finally, we remove those 3D Gaussians whose average distance $D > \mu +\sigma$ from the current segmentation results.

\begin{figure}[t]
\begin{minipage}[b]{0.31\linewidth}
  \centering
  \includegraphics[width=1.0\textwidth]{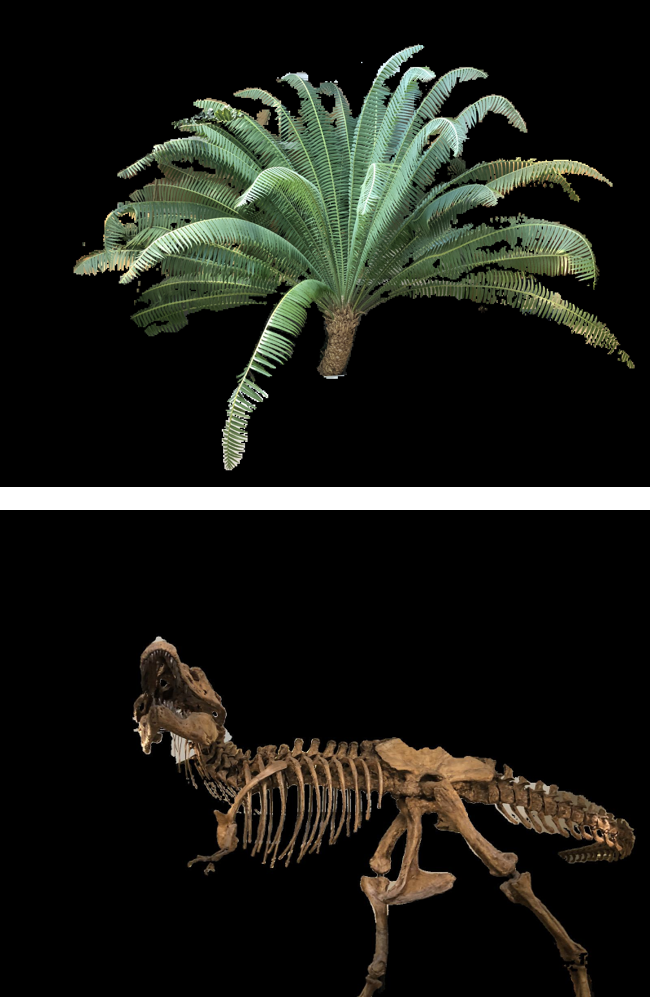}
% \centerline{\epsfig{figure=image3.ps,width=4.0cm}}
  \centerline{(a) GT Mask}\medskip
\end{minipage}
\hfill
\begin{minipage}[b]{0.31\linewidth}
  \centering
  \includegraphics[width=1.0\textwidth]{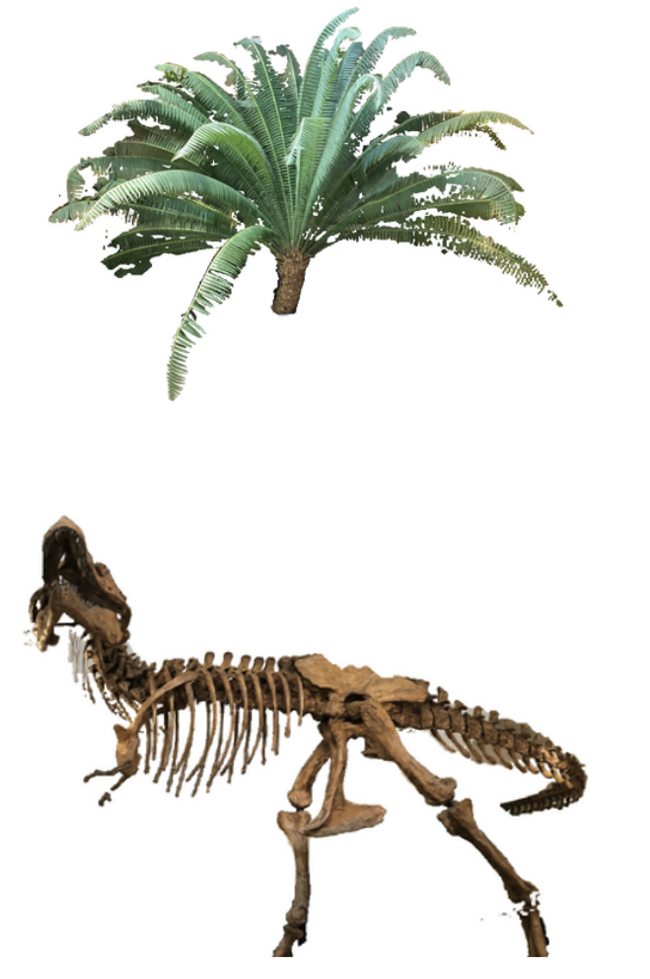}
% \centerline{\epsfig{figure=image4.ps,width=4.0cm}}
  \centerline{(b) ISRF result}\medskip
\end{minipage}
\hfill
\begin{minipage}[b]{0.31\linewidth}
  \centering
  \includegraphics[width=1.0\textwidth]{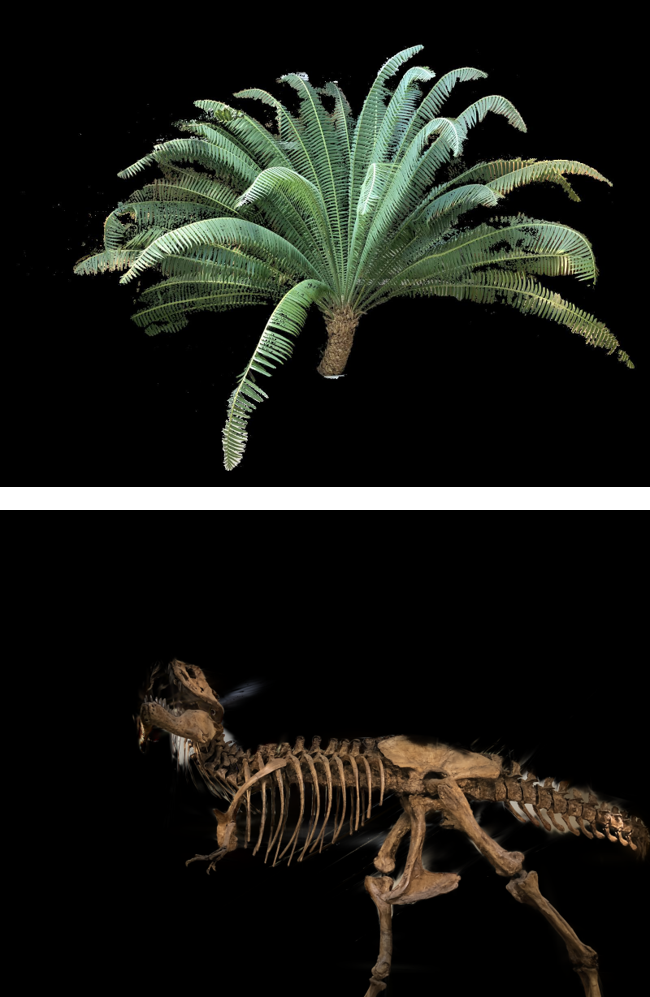}
% \centerline{\epsfig{figure=image4.ps,width=4.0cm}}
  \centerline{(c) our result}\medskip
\end{minipage}
\caption{Comparison results. (a) is the GT Mask we used to guide the segmentation of 3D Gaussians, (b) is the result of ISRF (from the original paper), and (c) is our result.}
\label{Comparison}
\end{figure}
\section{Experiment}
\subsection{Setups}
% 由于3D高斯分割方法的稀少以及最近的工作没有开源代码，我们选择将本方法与此前的NeRF分割方法作对比，并选取ISRF作为baseline。为此我们选取了常见的NeRF数据集进行实验，包括LLFF、NeRF-360以及Mip-NeRF 360。LLFF和NeRF-360均以场景中的物体为中心，区别在于前者相机视角在小范围内变化，后者包含了围绕物体的360°的图片。Mip-NeRF 360则包含了开放式的场景，其相机视角也在大范围内变化。在Gaussian Clustering阶段，每个3D高斯的概率阈值设置为0.65，同时筛选出与其距离最接近的50个3D高斯进行后续计算。实验中采用的优化器为Adam，初始学习率为0.05，采用指数衰减策略。
Due to the scarcity of 3D Gaussian segmentation methods and the lack of open source code for Gaussian Grouping~\cite{Gaussian_Grouping} and SAGA~\cite{SAGA}, we chose to compare our method with previous NeRF segmentation methods~\cite{ISRF}. For this purpose, we selected well-known NeRF datasets for our experiments, including LLFF~\cite{LLFF}, NeRF-360~\cite{NeRF}, and Mip-NeRF 360~\cite{MIP-NeRF-360}. Both LLFF and NeRF-360 are centered on objects in the scene, with the difference that the camera viewpoint of the former varies in a small range, while the latter contains a 360° image around the object. Mip-NeRF 360 features an unbounded scene, and its camera viewpoint also varies in a large range. In the Gaussian Clustering stage, the probability threshold $\beta$ of each 3D Gaussian is set at 0.65, while the 50 3D Gaussians closest to its distance are filtered for subsequent computation. The 3D Gaussians are built and trained on a single Nvidia Geforce RTX 3090 GPU.

\begin{figure}[t]
\begin{minipage}[b]{0.24\linewidth}
  \centering
  \includegraphics[width=1.0\textwidth]{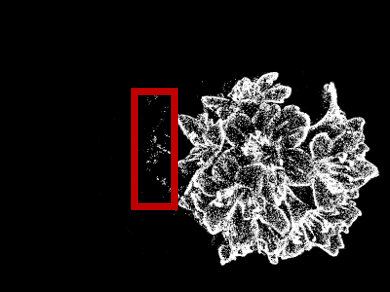}
% \centerline{\epsfig{figure=image3.ps,width=4.0cm}}
  \centerline{(a) Original points}\medskip
\end{minipage}
\hfill
\begin{minipage}[b]{0.24\linewidth}
  \centering
  \includegraphics[width=1.0\textwidth]{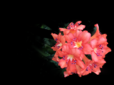}
% \centerline{\epsfig{figure=image3.ps,width=4.0cm}}
  \centerline{(b) Original}\medskip
\end{minipage}
\hfill
\begin{minipage}[b]{0.24\linewidth}
  \centering
  \includegraphics[width=1.0\textwidth]{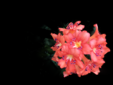}
% \centerline{\epsfig{figure=image4.ps,width=4.0cm}}
  \centerline{(c) KNN}\medskip
\end{minipage}
\hfill
\begin{minipage}[b]{0.24\linewidth}
  \centering
  \includegraphics[width=1.0\textwidth]{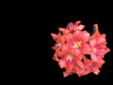}
% \centerline{\epsfig{figure=image4.ps,width=4.0cm}}
  \centerline{(d) KNN+Filter}\medskip
\end{minipage}
\caption{Ablation results. (a) is the segmented Gaussian center point, (b) is the original segmentation result without KNN or Filtering, (c) is the result after KNN, and (d) is the final result with KNN and Filtering}
\label{Ablation}
\end{figure}

\subsection{Result}

% 图二展示了本方法在不同场景下的分割效果。前两行展示了其在相机位姿在小范围内变动时的分割效果，第三行展示了其在360°场景下的分割效果。最后一行则给出了其在进行多物体分割时的效果，图中的电视、desk和table分别属于不同的物体。我们添加的object code作为一种简单高效的表达方式使得我们的方法可以处理一些复杂的场景。第一行中花后方繁杂的叶子被成功去除，第三行在观察视角变化了一个大的角度的情况下也取得了很好的分割结果。此外，该object code由于包含了3D高斯在所有类别上的概率分布信息，使得我们的方法可以同时分割出场景中的多个物体。
Fig. \ref{Experiment} illustrates the segmentation effects of this method in various scenes. The first two rows demonstrate the segmentation performance when the camera position varies within a small range. The third row depicts the segmentation effect in a 360° scene. The final row highlights the results of multi-object segmentation, where distinct objects such as the TV, desk, and table are segmented separately. Our method's efficiency is enhanced by the addition of object code, a simple yet effective tool for handling complex scenes. In the first row, this code enables the successful removal of complex background elements like the leaves behind the flower. In the third row, it ensures accurate segmentation even when there is a significant change in the viewing angle. Moreover, the object code, which encapsulates the probability distribution of the 3D Gaussian across all classes, facilitates the simultaneous segmentation of multiple objects in a scene.

% 图三展示了本方法与ISRF的对比效果。由于3D高斯的显式表征，我们的分割结果相比于ISRF在细节部分更加精确，这在图中的叶子部分尤其明显。
Fig. \ref{Comparison} illustrates the comparative results between our method and ISRF~\cite{ISRF}. Owing to the explicit representation of 3D Gaussians, our segmentation results are more precise in detail compared to those of ISRF, which is particularly evident in the leaf section of Fig. \ref{Comparison}. 

\subsection{Ablations}
% 图四给出了消融实验的结果，充分展现了KNN与统计过滤的有效性。图四（a）是初始的未经KNN或是统计过滤得到的分割前景图，可以看到花后方的一些叶子也被错误的分割出来。图四（b）则是经过KNN聚类后的分割前景图，由于KNN聚类主要处理那些语义信息模糊的高斯，所以其对可视化的结果影响不大，但也可以看出其去除了一些被错误分割出来的高斯。最后图四（c）则是同时经过KNN和统计过滤得到的结果，可以看到其成功的过滤了那些被错误分割出来的高斯。
Fig. \ref{Ablation} presents the results of the ablation experiments, clearly demonstrating the effectiveness of KNN clustering and statistical filtering. Fig. \ref{Ablation}(b) shows the initial segmented foreground object obtained without KNN clustering or statistical filtering, where it is noticeable that some leaves behind the flower are erroneously segmented. Fig. \ref{Ablation}(c) displays the segmented foreground image after KNN clustering. Since KNN primarily addresses Gaussians with ambiguous semantic information, its impact on the visualization result is minimal. However, it can be observed that some incorrectly segmented Gaussians have been removed. Finally, Fig. \ref{Ablation}(d) shows the result obtained after applying both KNN clustering and statistical filtering, which successfully filters out those Gaussians that were incorrectly segmented.

\section{Conclusion}
% 我们提出了一个使用2D分割图监督实现的3D高斯分割方法。为每个3D高斯添加的在各类别上的概率分布向量使得场景中的大部分3D高斯能被正确分割。同时KNN聚类利用了空间中物体的连续性来使得相邻的高斯有着相同的类别，可选的统计过滤帮助去除那些被错误分割出来的3D高斯。作为3D理解与编辑的第一步，本方法在下游任务中有着广泛的应用前景。
We propose a 3D Gaussian segmentation method guided by 2D segmentation maps, attaching a probability distribution vector for each 3D Gaussian on various categories to enable the segmentation of the majority of 3D Gaussians in the scene. Meanwhile, we employ KNN clustering to utilize the spatial continuity of objects, ensuring that nearby 3D Gaussians belong to the same category. Additionally, optional statistical filtering is used to help remove those 3D Gaussians that are incorrectly segmented. As an initial step in 3D understanding and editing, this method has a wide range of potential applications in downstream tasks. We demonstrate the effectiveness of our method on common NeRF datasets.

\bibliographystyle{IEEEbib}

\bibliography{icme2023template}

\begin{thebibliography}{10}

\bibitem{3D-Gaussian}
Bernhard Kerbl, Georgios Kopanas, Thomas Leimk{\"u}hler, and George Drettakis,
\newblock ``{3D Gaussian Splatting for Real-Time Radiance Field Rendering},''
\newblock {\em {ACM Transactions on Graphics}}, vol. 42, no. 4, pp. 1--14, July 2023.

\bibitem{point-cloud}
Weiping Liu, Jia Sun, Wanyi Li, Ting Hu, and Peng Wang,
\newblock ``Deep learning on point clouds and its application: A survey,''
\newblock {\em Sensors}, vol. 19, no. 19, 2019.

\bibitem{mesh}
Dawar Khan, Alexander Plopski, Yuichiro Fujimoto, Masayuki Kanbara, Gul Jabeen, Yongjie~Jessica Zhang, Xiaopeng Zhang, and Hirokazu Kato,
\newblock ``Surface remeshing: A systematic literature review of methods and research directions,''
\newblock {\em IEEE Transactions on Visualization and Computer Graphics}, vol. 28, no. 3, pp. 1680--1713, 2022.

\bibitem{SDF}
Jeong~Joon Park, Peter Florence, Julian Straub, Richard Newcombe, and Steven Lovegrove,
\newblock ``Deepsdf: Learning continuous signed distance functions for shape representation,''
\newblock in {\em Proceedings of the IEEE/CVF Conference on Computer Vision and Pattern Recognition (CVPR)}, June 2019.

\bibitem{NeRF}
Ben Mildenhall, Pratul~P. Srinivasan, Matthew Tancik, Jonathan~T. Barron, Ravi Ramamoorthi, and Ren Ng,
\newblock ``Nerf: Representing scenes as neural radiance fields for view synthesis,''
\newblock {\em Commun. ACM}, vol. 65, no. 1, pp. 99–106, dec 2021.

\bibitem{SFM}
Johannes~L. Schonberger and Jan-Michael Frahm,
\newblock ``Structure-from-motion revisited,''
\newblock in {\em Proceedings of the IEEE Conference on Computer Vision and Pattern Recognition (CVPR)}, June 2016.

\bibitem{Gaussian_Grouping}
Mingqiao {Ye}, Martin {Danelljan}, Fisher {Yu}, and Lei {Ke},
\newblock ``{Gaussian Grouping: Segment and Edit Anything in 3D Scenes},''
\newblock {\em arXiv e-prints}, p. arXiv:2312.00732, Dec. 2023.

\bibitem{SAGA}
Jiazhong {Cen}, Jiemin {Fang}, Chen {Yang}, Lingxi {Xie}, Xiaopeng {Zhang}, Wei {Shen}, and Qi~{Tian},
\newblock ``{Segment Any 3D Gaussians},''
\newblock {\em arXiv e-prints}, p. arXiv:2312.00860, Dec. 2023.

\bibitem{Decomposing}
Sosuke Kobayashi, Eiichi Matsumoto, and Vincent Sitzmann,
\newblock ``Decomposing nerf for editing via feature field distillation,''
\newblock in {\em Advances in Neural Information Processing Systems}, S.~Koyejo, S.~Mohamed, A.~Agarwal, D.~Belgrave, K.~Cho, and A.~Oh, Eds. 2022, vol.~35, pp. 23311--23330, Curran Associates, Inc.

\bibitem{ISRF}
Rahul Goel, Dhawal Sirikonda, Saurabh Saini, and P.~J. Narayanan,
\newblock ``Interactive segmentation of radiance fields,''
\newblock in {\em Proceedings of the IEEE/CVF Conference on Computer Vision and Pattern Recognition (CVPR)}, June 2023, pp. 4201--4211.

\bibitem{DreanGaussian}
Jiaxiang {Tang}, Jiawei {Ren}, Hang {Zhou}, Ziwei {Liu}, and Gang {Zeng},
\newblock ``{DreamGaussian: Generative Gaussian Splatting for Efficient 3D Content Creation},''
\newblock {\em arXiv e-prints}, p. arXiv:2309.16653, Sept. 2023.

\bibitem{GaussianDreamer}
Taoran {Yi}, Jiemin {Fang}, Junjie {Wang}, Guanjun {Wu}, Lingxi {Xie}, Xiaopeng {Zhang}, Wenyu {Liu}, Qi~{Tian}, and Xinggang {Wang},
\newblock ``{GaussianDreamer: Fast Generation from Text to 3D Gaussians by Bridging 2D and 3D Diffusion Models},''
\newblock {\em arXiv e-prints}, p. arXiv:2310.08529, Oct. 2023.

\bibitem{Diffusion}
Robin Rombach, Andreas Blattmann, Dominik Lorenz, Patrick Esser, and Bj\"orn Ommer,
\newblock ``High-resolution image synthesis with latent diffusion models,''
\newblock in {\em Proceedings of the IEEE/CVF Conference on Computer Vision and Pattern Recognition (CVPR)}, June 2022, pp. 10684--10695.

\bibitem{4D}
Guanjun {Wu}, Taoran {Yi}, Jiemin {Fang}, Lingxi {Xie}, Xiaopeng {Zhang}, Wei {Wei}, Wenyu {Liu}, Qi~{Tian}, and Xinggang {Wang},
\newblock ``{4D Gaussian Splatting for Real-Time Dynamic Scene Rendering},''
\newblock {\em arXiv e-prints}, p. arXiv:2310.08528, Oct. 2023.

\bibitem{SAM}
Alexander {Kirillov}, Eric {Mintun}, Nikhila {Ravi}, Hanzi {Mao}, Chloe {Rolland}, Laura {Gustafson}, Tete {Xiao}, Spencer {Whitehead}, Alexander~C. {Berg}, Wan-Yen {Lo}, Piotr {Doll{\'a}r}, and Ross {Girshick},
\newblock ``{Segment Anything},''
\newblock {\em arXiv e-prints}, p. arXiv:2304.02643, Apr. 2023.

\bibitem{DM-NeRF}
Bing {Wang}, Lu~{Chen}, and Bo~{Yang},
\newblock ``{DM-NeRF: 3D Scene Geometry Decomposition and Manipulation from 2D Images},''
\newblock {\em arXiv e-prints}, p. arXiv:2208.07227, Aug. 2022.

\bibitem{Object-NeRF}
Bangbang Yang, Yinda Zhang, Yinghao Xu, Yijin Li, Han Zhou, Hujun Bao, Guofeng Zhang, and Zhaopeng Cui,
\newblock ``Learning object-compositional neural radiance field for editable scene rendering,''
\newblock in {\em Proceedings of the IEEE/CVF International Conference on Computer Vision (ICCV)}, October 2021, pp. 13779--13788.

\bibitem{OR-NeRF}
Youtan {Yin}, Zhoujie {Fu}, Fan {Yang}, and Guosheng {Lin},
\newblock ``{OR-NeRF: Object Removing from 3D Scenes Guided by Multiview Segmentation with Neural Radiance Fields},''
\newblock {\em arXiv e-prints}, p. arXiv:2305.10503, May 2023.

\bibitem{SegNeRF}
Jesus {Zarzar}, Sara {Rojas}, Silvio {Giancola}, and Bernard {Ghanem},
\newblock ``{SegNeRF: 3D Part Segmentation with Neural Radiance Fields},''
\newblock {\em arXiv e-prints}, p. arXiv:2211.11215, Nov. 2022.

\bibitem{SPIn-NeRF}
Ashkan Mirzaei, Tristan Aumentado-Armstrong, Konstantinos~G. Derpanis, Jonathan Kelly, Marcus~A. Brubaker, Igor Gilitschenski, and Alex Levinshtein,
\newblock ``Spin-nerf: Multiview segmentation and perceptual inpainting with neural radiance fields,''
\newblock in {\em Proceedings of the IEEE/CVF Conference on Computer Vision and Pattern Recognition (CVPR)}, June 2023, pp. 20669--20679.

\bibitem{Switch-NeRF}
Zhenxing MI and Dan Xu,
\newblock ``Switch-ne{RF}: Learning scene decomposition with mixture of experts for large-scale neural radiance fields,''
\newblock in {\em The Eleventh International Conference on Learning Representations}, 2023.

\bibitem{DINO}
Mathilde Caron, Hugo Touvron, Ishan Misra, Herv\'e J\'egou, Julien Mairal, Piotr Bojanowski, and Armand Joulin,
\newblock ``Emerging properties in self-supervised vision transformers,''
\newblock in {\em Proceedings of the IEEE/CVF International Conference on Computer Vision (ICCV)}, October 2021, pp. 9650--9660.

\bibitem{eiseg}
Yuying Hao, Yi~Liu, Yizhou Chen, Lin Han, Juncai Peng, Shiyu Tang, Guowei Chen, Zewu Wu, Zeyu Chen, and Baohua Lai,
\newblock ``Eiseg: An efficient interactive segmentation tool based on paddlepaddle,''
\newblock {\em arXiv e-prints}, pp. arXiv--2210, 2022.

\bibitem{LLFF}
Ben Mildenhall, Pratul~P. Srinivasan, Rodrigo Ortiz-Cayon, Nima~Khademi Kalantari, Ravi Ramamoorthi, Ren Ng, and Abhishek Kar,
\newblock ``Local light field fusion: Practical view synthesis with prescriptive sampling guidelines,''
\newblock {\em ACM Trans. Graph.}, vol. 38, no. 4, jul 2019.

\bibitem{MIP-NeRF-360}
Jonathan~T. Barron, Ben Mildenhall, Dor Verbin, Pratul~P. Srinivasan, and Peter Hedman,
\newblock ``Mip-nerf 360: Unbounded anti-aliased neural radiance fields,''
\newblock in {\em Proceedings of the IEEE/CVF Conference on Computer Vision and Pattern Recognition (CVPR)}, June 2022, pp. 5470--5479.

\end{thebibliography}

\end{document}